\documentclass[accepted]{uai2023} 

\usepackage[british]{babel}

\usepackage{natbib} 
\usepackage{mathtools} 
\usepackage{booktabs} 
\usepackage{tikz} 

\usepackage{mathtools}
\usepackage{amssymb}
\usepackage{amsmath}
\usepackage{amsfonts}
\usepackage{amsthm}
\usepackage{commath}
\usepackage{bm}
\usepackage{physics}
\usepackage{multicol}
\usepackage{bbm}
\usepackage{scalerel}
\usepackage{comment}
\usepackage[all=normal, mathspacing=tight, floats=tight, paragraphs=tight]{savetrees}
\usepackage{multirow, booktabs}
\usepackage[bf]{caption}
\usepackage{subcaption}
\usepackage{makecell}
\usepackage{graphicx}
\usepackage{stfloats}

\DeclarePairedDelimiterX{\infdivx}[2]{(}{)}{%
	#1\;\delimsize\|\;#2%
}

\newcommand{\tP}{{\tt P}}

\newcommand{\tp}{{\tt p}}
\newcommand{\tq}{{\tt q}}
\newcommand{\tpost}{{\tt p}(\bm\theta\vert\mathcal{D})}

\newcommand{\KL}{{\tt KL}\infdivx}
\newcommand{\Softmax}{{\tt Softmax}}

\newcommand{\LogSumExp}{{\tt LogSumExp}}
\newcommand{\Dir}{{\tt Dir}}

\DeclareMathOperator*{\argmax}{arg\,max}

\newcommand{\tpm}{\tiny{$\pm$} }

\usepackage{pifont}
\newcommand{\xmark}{\ding{55}}%

\title{Logit-Based Ensemble Distribution Distillation for \\Robust Autoregressive Sequence Uncertainties}

\author[1]{\href{mailto:<yf286@cam.ac.uk>?Subject=L-EDD UAI 2023}{Yassir~Fathullah}{}}
\author[2]{\href{mailto:<g.xia21@imperial.ac.uk}{Guoxuan~Xia}{}}
\author[1]{Mark~J.~F.~Gales}
\affil[1]{%
    Engineering Department\\
    University of Cambridge\\
    UK
}
\affil[2]{%
    Department of Electrical \& Electronic Engineering\\
    Imperial College London\\
    UK
}
  
\begin{document}
\maketitle

\begin{abstract}

    Efficiently and reliably estimating uncertainty is an important objective in deep learning. It is especially pertinent to autoregressive sequence tasks, where training and inference costs are typically very high. However, existing research has predominantly focused on tasks with static data such as image classification. 
    In this work, we investigate Ensemble Distribution Distillation (EDD) applied to large-scale natural language sequence-to-sequence data. EDD aims to compress the superior uncertainty performance of an expensive (teacher) ensemble into a cheaper (student) single model. Importantly, the ability to separate knowledge (epistemic) and data (aleatoric) uncertainty is retained. Existing probability-space approaches to EDD, however, are difficult to scale to large vocabularies. 
    We show, for modern transformer architectures on large-scale translation tasks, that modelling the ensemble \textit{logits}, instead of softmax probabilities, leads to significantly better students. Moreover, the students surprisingly even \textit{outperform Deep Ensembles} by up to $\sim$10\% AUROC on out-of-distribution detection, whilst matching them at in-distribution translation. 

\end{abstract}

\section{Introduction}
\label{sec:intro}

The ability to produce reliable estimates of uncertainty is important to many tasks in deep learning. When it is costly to make mistakes, a model should know when to discard a prediction or defer it to a human expert. Although there is a significant body of research in this area \citep{ovadia_trust,duku,yang2021oodsurvey}, it tends to focus on \textit{static} data, where outputs are of a fixed dimension, with approaches most commonly evaluated on image classification \citep{pmlr-v97-geifman19a, sngp, yang2022openood,  Moon2020ConfidenceAwareLF, Xia_2022_ACCV}. In contrast, in this work, we aim to investigate uncertainty estimation for sequence prediction tasks, such as machine translation, which is a relatively under-explored domain \citep{structured}.

Large attention-based autoregressive neural networks have recently emerged as the most competitive approach to many structured sequence-prediction tasks, especially in translation \citep{atten, aiayn, scaling-nmt}, and are increasingly being used in practice. However, as the computational and memory costs of these modern approaches are typically very large, it is particularly important that approaches for improving the quality of uncertainties should be \textit{efficient}. We will focus on one such efficient approach to better uncertainties, Ensemble Distribution Distillation (EDD) \citep{endd}.

Ensembling multiple neural networks trained using different random seeds is a well-established approach for boosting uncertainty performance \citep{deepens}. Deep Ensembles have been shown to be effective over a wide range of data, tasks, and evaluation metrics \citep{ovadia_trust, structured, Kim2021AUB, gustafsson2020evaluating}. Moreover, they are naturally able to decompose total uncertainty into knowledge (epistemic) and data (aleatoric) uncertainty \citep{duku}, which can be useful for different tasks such as active learning \citep{pmlr-v70-gal17a, seq-al2}, reinforcement learning \citep{decomp}, and out-of-distribution detection \citep{structured}. However, Deep Ensembles suffer from costs that scale linearly with the number of members. EDD aims to tackle this by using Knowledge Distillation (KD) \citep{kd} to compress the (teacher) ensemble into a more efficient (student) single model. Crucially, EDD not only has the student learn the predictions of the ensemble but also the \textit{distribution} over individual ensemble member outputs. By explicitly modelling the diversity over the ensemble, the student is able to express knowledge and data uncertainty independently just like the teacher ensemble \citep{endd}.

However, EDD is not without its challenges. Prior work has shown that EDD suffers from optimisation issues, meaning it can be difficult to scale to confident ensembles with large label spaces. Thus, EDD requires a number of practical modifications in order to be applied to large-scale tasks such as machine translation \citep{seq-endd, ryabin}. Despite these challenges, the concept behind EDD remains a promising approach for training single autoregressive models with smaller footprints and the ability to estimate high-quality, robust uncertainties.

\textbf{Summary of contributions:} In this paper, we focus on an underexplored area of uncertainty estimation: robust and efficient autoregressive sequence uncertainties. Specifically, we address the drawbacks of sequence EDD by using \textit{logit-based} ensemble distribution distillation (L-EDD). Instead of training a student to distribution distil the information from an ensemble in the probability/softmax space, we teach it to perform the same task in the pre-softmax logit space. Experiments on the En-De WMT'16 and En-Ru WMT'20 machine translation tasks show that L-EDD, in particular when using a Laplace distribution, produces strong estimates of sequence uncertainty. L-EDD is able to outperform EDD and surprisingly even Deep Ensembles on out-of-distribution (OOD) detection and match them for translation quality.
%
Furthermore, by using Snapshot Ensembles \citep{snapshot}, we are able to greatly reduce the overall training costs of EDD compared to using a Deep Ensemble teacher.

\section{Background}
\label{sec:background}

In this section, we review ensemble-based uncertainty estimation. We follow with a discussion of how the limitations of ensembles can be addressed using recently developed distillation techniques for autoregressive sequence tasks such as machine translation. 

\subsection{Uncertainty Estimation}
\label{ssec:uncertainty}

We adopt a Bayesian perspective on ensembles as this offers a flexible framework within which uncertainties have an information-theoretic justification. The posterior over model parameters $\tpost$ is derived given some observed (training) data, ${\cal D}$. Unfortunately, the posterior is often intractable and cannot be derived for large non-linear networks. Alternatively an approximation $\tq(\bm\theta) \approx \tpost$ can be used. Samples from this approximate distribution can then be drawn to generate an ensemble of models. 

Take an ensemble $\{\tP(\bm y\vert\bm x, \bm\theta^{(m)})\}_{m = 1}^{M}$ sampled from an approximate posterior $\tq(\bm\theta)$ where each model maps a \textit{variable-length} input $\bm x \in \mathcal{X}$ into a \textit{variable-length} output $\bm y \in \mathcal{Y}$ of discrete units. The predictive distribution is obtained by:
\begin{align}
	\label{eq:predictive}
	\tP(\bm y\vert\bm x, \mathcal{D}) = \mathbb{E}_{\tq(\bm\theta)} \left[ \tP(\bm y\vert\bm x, \bm\theta) \right].
\end{align}
From this predictive distribution, a measure of total uncertainty can be estimated using the entropy:
%
\begin{align}
	\label{eq:total}
	\mathcal{H}\left[ \tP(\bm y\vert\bm x, \mathcal{D}) \right] = \mathbb{E}_{\tP(\bm y\vert\bm x, \mathcal{D})}\left[ -\ln \tP(\bm y\vert\bm x, \mathcal{D}) \right].
\end{align}
Furthermore, a measure of disagreement between models, also referred to as \textit{knowledge} or \textit{epistemic} uncertainty, can be estimated by using mutual information between $\bm y$ and $\bm\theta$:
\begin{align}
	\label{eq:knowledge}
	\begin{split}
		\mathcal{I}\left[ \bm y, \bm\theta \vert \bm x, \mathcal{D} \right] = \mathbb{E}_{\tq(\bm\theta)}\left[ \KL[\big]{\tP(\bm y\vert\bm x, \bm\theta)}{\tP(\bm y\vert\bm x, \mathcal{D})} \right].
	\end{split}
\end{align}
This estimate can also be decomposed into a measure of total and data (aleatoric) uncertainty, as mentioned in \citet{structured}. There are also many other potential measures of knowledge uncertainty such as expected pairwise KL-divergence or reverse mutual information \citep{structured}, however, for the sake of simplicity we restrict our focus to the already mentioned eq. (\ref{eq:total}) and (\ref{eq:knowledge}) since these represent uncertainties of differing natures.

\textbf{Limitations}: The discussion has so far assumed one can enumerate all possible variable-length outputs $\bm y \in \mathcal{Y}$ which is not tractable in autoregressive sequence tasks. Instead, one can approximate the uncertainties by monte-carlo methods \citep{notin2021improving} and utilising the autoregressive structure of predictions \citep{structured}:
\begin{align}
	\label{eq:ar}
	\begin{split}
		\tP(\bm y\vert\bm x, \bm\theta) = \prod_{l = 1}^{L} \tP(y_l \vert \bm y_{<l}, \bm x, \bm\theta).
	\end{split}
\end{align}
We refer to \citet{structured} for an in-depth discussion and analysis of approximations for predictive entropy and mutual information for autoregressive prediction.

\subsection{Knowledge Distillation}
\label{ssec:distillation}

Ensembles $\{\tP(\bm y\vert\bm x, \bm\theta^{(m)})\}_{m = 1}^{M}$ sampled from some posterior can be computationally demanding. One approach to efficiently exploit the information of the ensemble is to use Knowledge Distillation (KD) to yield a single student model \citep{kd, seq-kd}. 

Given a reference data pair $(\bm x, \bm y) \sim \tilde{\tp}(\bm x, \bm y)$, a standard model might be trained using negative log-likelihood (NLL): 
\begin{align}
	\label{eq:nll}
	\mathcal{L}_{\tt NLL}(\bm\theta) = - \frac{1}{L} \sum_{l = 1}^{L} \ln\tP(y_l \vert \bm y_{<l}, \bm x, \bm\theta).
\end{align}
This is referred to as teacher-forcing since during training the model makes predictions at step $l$ conditioned on the true output $\bm y_{<l}$ (rather than its own previous predictions). Similarly, a student model with parameters $\bm\phi$ can be trained to emulate a teacher ensemble by additionally using \textit{average} ensemble categorical/softmax outputs $\bm\pi_l, \pi_{l, k} = \tP(y_l = k \vert \bm y_{<l}, \bm x, \mathcal{D})$ as soft labels:
\begin{align}
	\label{eq:distillation}
	\begin{split}
		\mathcal{L}_{\tt KL}(\bm\phi) = \frac{1}{L}\sum_{l = 1}^{L} \KL[\big]{\bm\pi_l}{\tP(y_l \vert \bm y_{<l}, \bm x, \bm\phi)}.
	\end{split}
\end{align}
However in practice, one optimises a convex combination of the likelihood and KL-divergence losses $\mathcal{L}_{\tt KD}(\bm\phi) = \lambda \mathcal{L}_{\tt NLL}(\bm\phi) + (1-\lambda)\mathcal{L}_{\tt KL}(\bm\phi), \medspace \lambda \in [0, 1]$ for added supervision and stability. The probability mass functions in the KL-divergence can also be temperature scaled to improve optimisation \citep{kd}. Note that this criterion is only considered for the teacher-forcing case, more sophisticated distillation approaches exist, by sampling $(\bm x, \bm y)$ from alternative distributions, but is outside the scope for this work, see \citet{seq-kd, gen-student-th, ood-training-1, ood-training-2} for details.

\subsection{Ensemble Distribution Distillation}
\label{ssec:endd}

Whilst KD has been successful in many sequence tasks, the resulting student is not able to estimate knowledge uncertainty, since it only models the average ensemble output. To avoid this issue, \citet{endd, ryabin} consider the task of distilling the \textit{distribution} of sequence ensemble predictions onto a single student. This allows the student to retain both predictive performance and information about ensemble diversity. 

To explain the mechanics behind Ensemble Distribution Distillation (EDD), consider modelling a distribution over autoregressive ensemble predictions, in which all $M$ ensemble members share the same back-history $\bm y_{<l}$:
\begin{align}
    \label{eq:predictionsens}
    \{\bm\pi_{l}^{(m)}\}_{m=1}^M, \medspace \pi_{l, k}^{(m)} = \tP(y_l = \omega_k\vert\bm y_{<l}, \bm x, \bm\theta^{(m)}).
\end{align}
Now let an autoregressive student predict the parameters $\bm\alpha_l$ of a Dirichlet distribution $\Dir(\bm\pi_l\vert\bm\alpha_l) = \tp(\bm\pi_l\vert\bm y_{<l}, \bm x, \bm\phi)$. Since the Dirichlet models
a distribution over categorical distributions it is an ideal candidate for this task. The distribution distillation loss of such a model is then simply the result of (negative) log-likelihood:
\begin{align}
    \label{eq:dirichletdd}
    \begin{split}
        \mathcal{L}_{\tt NLL}^{\tt DD}(\bm\phi) 
        & = -\frac{1}{MLK} \sum_{m, l} \ln\Dir(\bm \pi_l^{(m)}\vert\bm\alpha_l) \\
        & \equiv \frac{1}{LK} \sum_{l} \Big( \ln B(\bm\alpha_l) - \sum_{k} \alpha_{l, k}\ln\tilde{\pi}_{l, k} \Big), \
    \end{split}
\end{align}
where $K$ is the number of classes, $B(\bm\alpha)$ is the beta function and $\tilde{\pi}_{l, k}$ is the geometric average of the individual ensemble softmax probabilities in Equation (\ref{eq:predictionsens}). 

Whilst this approach was shown to be promising on a small-scale image classification task in \cite{endd}, following work \citep{seq-endd, ryabin} found that direct application of Equation (\ref{eq:dirichletdd}) encounters optimisation issues when scaled to larger label spaces. 
This arises from the way classwise loss gradients are related to teacher class probabilities.
It turns out that, unlike standard distillation, the loss in Equation (\ref{eq:dirichletdd}) induces small gradients for (important) high-probability classes and large gradients for (unimportant) low-probability classes. This negatively affects convergence as the number of low-probability classes increases.
\citet{ryabin} proposed an approach for scaling Dirichlet EDD, where the student aims to minimise a normalized reverse KL-divergence to a \textit{proxy} Dirichlet, which will be used as a baseline in this work.

\section{Sequence Logit-based EDD}
\label{sec:ledd}

In sequence tasks with a large number of classes, which commonly occurs in speech recognition and machine translation, the output categorical distributions are often very sparse and concentrated. Therefore, it often becomes highly challenging to apply EDD to tasks of this nature. On the other hand, KD has been shown to work well for larger tasks \citep{seq-kd, seq-kd-app1, seq-kd-app2, seq-kd-app3}, but since it only models the average teacher predictions, it cannot estimate data and knowledge uncertainties that are important for many downstream tasks such as out-of-distribution detection.

In this section, we describe a \textit{Logit-based} Ensemble Distribution Distillation (L-EDD) approach for autoregressive models which addresses the drawbacks of both KD and EDD in a single consistent framework and is scalable to sequence problems with a large number of classes. Consider a set of logits produced by an ensemble:
\begin{align}
    \label{eq:logits}
    \{\bm z_{l}^{(m)}\}_{m=1}^M, \medspace \bm\pi_{l}^{(m)} = \Softmax(\bm z_{l}^{(m)}).
\end{align}
Traditional distillation approaches thereafter use the logits to produce categorical probability distributions by applying the softmax function. However, instead of operating in the probability space, we propose training a student, with model parameters $\bm\phi$, to directly model the logit space by predicting the mean $\bm\mu_l$ and scale $\bm\sigma_l$ parameters of a diagonal Laplace distribution:
\begin{align}
    \label{eq:laplace-logit}
    \begin{split}
        \tp(\bm z \vert \bm y_{<l}, \bm x, \bm\phi) 
        & = {\tt Lap}(\bm z \vert \bm\mu_{l}, \bm\sigma_{l}) \\
        & = \prod_{k} \frac{1}{2\sigma_{l,k}} \exp{-\frac{\abs{z_k - \mu_{l,k}}}{\sigma_{l,k}}}.
    \end{split}
\end{align}
Because we opt for a diagonal distribution, sampling is parallelisable, highly efficient, and straightforward and allows for the estimation of uncertainties in exactly the same manner as in standard ensembles. Additionally, significantly fewer parameters are required compared to using a fully-specified covariance matrix. Another benefit of the chosen distribution is the long tails which make the Laplace robust to outliers, unlike the Gaussian distribution. This robustness also makes it a natural choice for handling the early stages of training when the student model is randomly initialised and its output distribution substantially differs from the ensemble logits.

Furthermore, given the set of logits produced by an ensemble, the student model $\tp(\bm z \vert \bm y_{<l}, \bm x, \bm\phi)$ can be trained by straightforward application of log-likelihood training:
\begin{align}
    \label{eq:laplacedd}
    \begin{split}
        \mathcal{L}_{\tt NLL}^{\tt L-EDD}(\bm\phi) 
        & = -\frac{1}{MLK} \sum_{m, l} \ln{\tt Lap}(\bm z_l^{(m)} \vert \bm\mu_{l}, \bm\sigma_{l}) \\
        & \equiv \frac{1}{MLK} \sum_{m,l,k} \frac{\abs{z_{l,k}^{(m)} - \mu_{l,k}}}{\sigma_{l,k}} + \ln\sigma_{l,k}.
    \end{split}
\end{align}

We also perform experiments with a student (diagonal) Gaussian distribution on the logits, variations of which have been explored in static image classification \citep{s2d, mod-edd} but remained unexplored for autoregressive sequence tasks:
\begin{align}
    \label{eq:gaussian-logit}
    \begin{split}
        \tp(\bm z \vert \bm y_{<l}, & \bm x, \bm\phi) 
        = \mathcal{N}(\bm z \vert \bm\mu_{l}, \bm\sigma_{l}^2) \\
        & = \prod_{k} \frac{1}{(2\pi\sigma_{l,k}^2)^{\frac{1}{2}}} \exp{-\frac{(z_k - \mu_{l,k})^2}{2\sigma_{l,k}^2}}.
    \end{split}
\end{align}
Similar to all of the mentioned approaches, this system is also trained using the log-likelihood objective:
\begin{align}
    \label{eq:gaussiandd}
    \begin{split}
        \mathcal{L}_{\tt NLL}^{\tt L-EDD}(\bm\phi) 
        & = -\frac{1}{MLK} \sum_{m, l} \ln\mathcal{N}(\bm z_l^{(m)} \vert \bm\mu_{l}, \bm\sigma_{l}^2). 
    \end{split}
\end{align}
The Gaussian distribution, which induces an L2-norm loss function is much more sensitive to outliers in the ensemble outputs. This student could potentially be more challenging to train, but should still be more stable than Dirichlet EDD.

\subsection{Practical Considerations}

Since the softmax activation function is shift invariant,
\begin{align*}
    \Softmax(\bm z - \bm 1 b) = \Softmax(\bm z) \thickspace\thickspace \forall b \in \mathbb{R},
\end{align*}
one has to consider this property when performing distribution distillation. Ensemble members are unconstrained along $\bm 1$, and so can potentially vary wildly in the logit space, even if they give consistent softmax predictions. Therefore, logits are normalised by $\tilde{\bm z} = \bm z - \bm 1 \LogSumExp(\bm z)$. This particular normalisation scheme is not special and any choice of the normalisation constant such as ${\tt Max}(\bm z)$ or ${\tt Mean}(\bm z)$ would be valid. Next, to ensure that the student can be trained reliably, we interpolate the knowledge and distribution distillation losses $\mathcal{L}_{\tt KD}(\bm\phi) + \beta \mathcal{L}_{\tt NLL}^{\tt L-EDD}(\bm\phi)$ (see Eq. \ref{eq:laplacedd} and Sec. \ref{ssec:distillation}).

Furthermore, distributions in logit space often lead to analytically intractable expectations in probability space. The standard approach to circumvent this issue is by sampling from the distribution using monte-carlo approximations. However, in this paper, we opt for an approximative deterministic approach when computing the predictive distribution (e.g. when decoding):
\begin{align*}
    \tP(y_l \vert \bm y_{<l}, \bm x, \bm\phi) 
    & = \mathbb{E}_{\tp(\bm z_l \vert \bm y_{<l}, \bm x, \bm\phi)} \left[ \Softmax(\bm z_l) \right] \\
    & \approx \Softmax(\bm \mu_l),
\end{align*}
in which we approximate the expectation by just using the mean of the logit distribution. When performing downstream tasks that require uncertainties we revert to a stochastic sampling scheme to generate multiple predictions from the distribution.

\section{Experiments on Artificial Data}
\label{sec:toy-example}

\begin{figure*}[t!]
     \centering
             \begin{subfigure}{0.24\textwidth}
                 \centering
                 \vstretch{1.0}{\includegraphics[width=\textwidth]{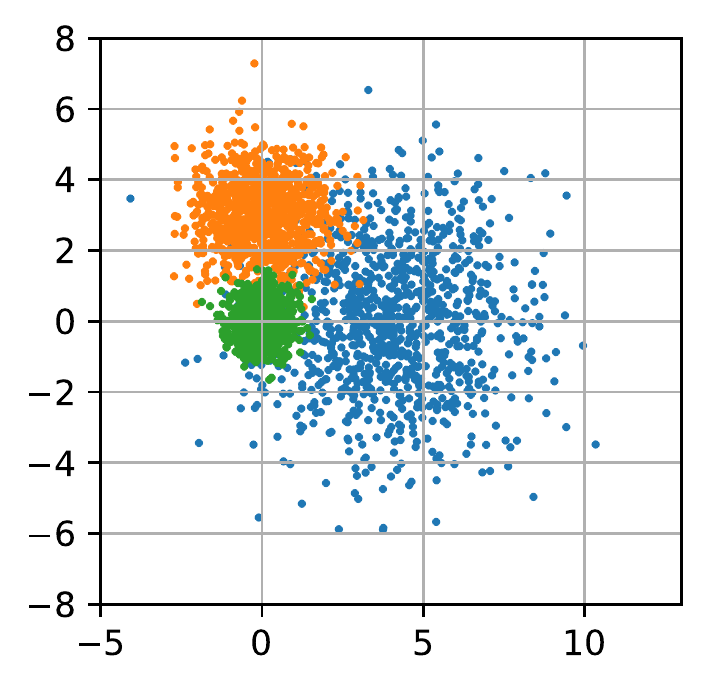}}
                 
                 \caption{Dataset.}
                 \label{fig:toy-data}
                 
             \end{subfigure}
             \hfill
             \begin{subfigure}{0.24\textwidth}
                 \centering
                 \vstretch{1.0}{\includegraphics[width=\textwidth]{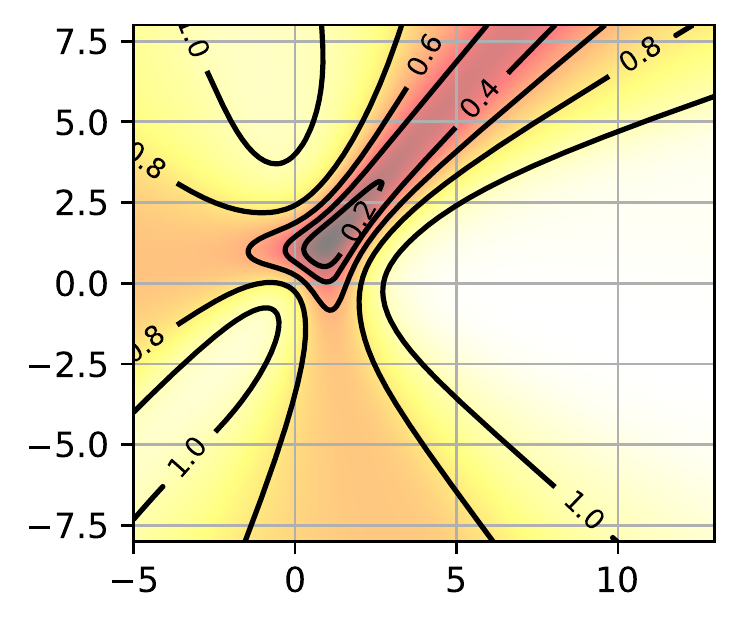}}
                 
                 \caption{Distillation loss}
                 \label{fig:toy-dist-loss}
                 
             \end{subfigure}
             \hfill
             \begin{subfigure}{0.24\textwidth}
                 \centering
                 \vstretch{1.0}{\includegraphics[width=\textwidth]{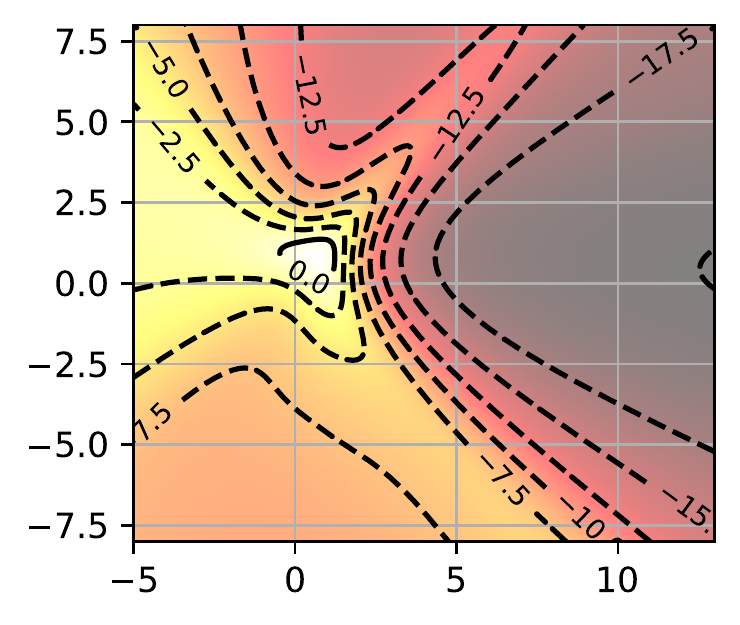}}
                 
                 \caption{Dirichlet loss}
                 \label{fig:toy-dir-loss}
                 
             \end{subfigure}
             \hfill
             \begin{subfigure}{0.24\textwidth}
                 \centering
                 \vstretch{1.0}{\includegraphics[width=\textwidth]{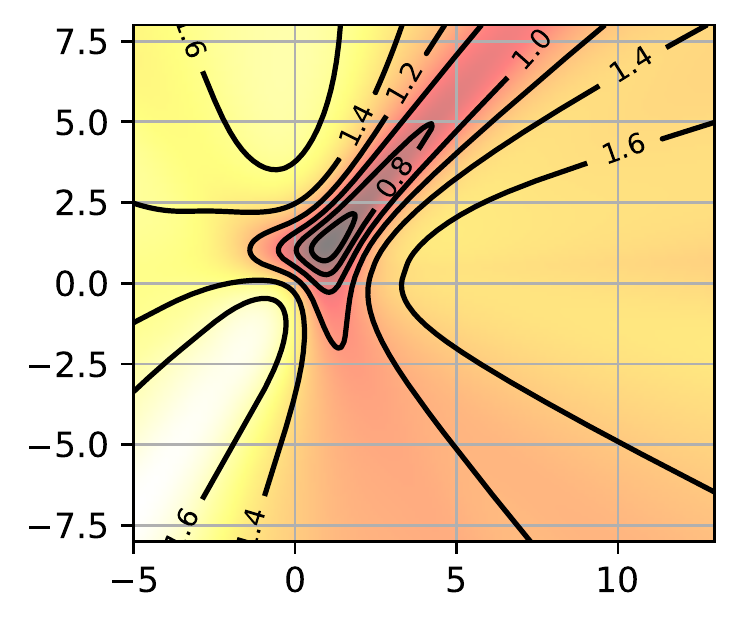}}
                 
                 \caption{Laplace loss}
                 \label{fig:toy-lap-loss}
                 
             \end{subfigure}
             
             \begin{subfigure}{0.24\textwidth}
                 \centering
                 \vstretch{1.0}{\includegraphics[width=\textwidth]{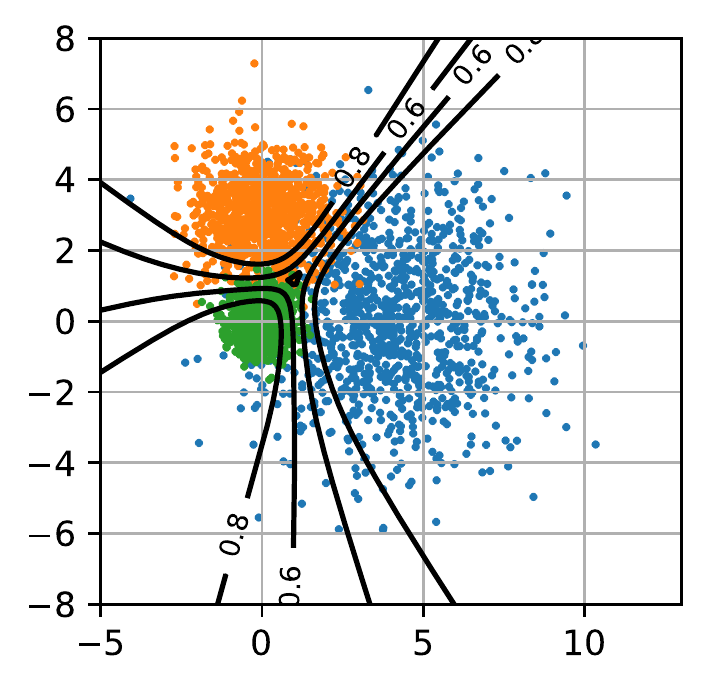}}
                 
                 \caption{Ensemble conf.}
                 \label{fig:toy-ens-conf}
             \end{subfigure}
             \hfill
             \begin{subfigure}{0.24\textwidth}
                 \centering
                 \vstretch{1.0}{\includegraphics[width=\textwidth]{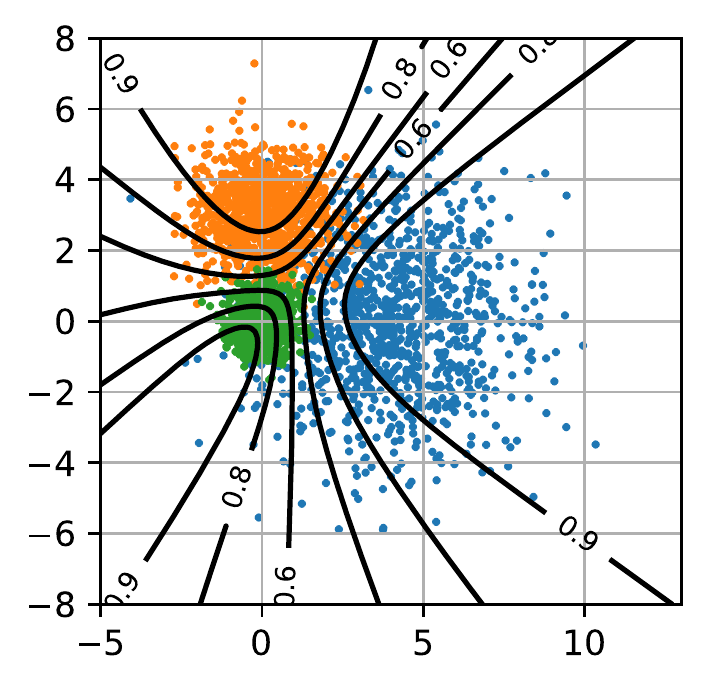}}
                 
                 \caption{Distillation conf.}
                 \label{fig:toy-dist-conf}
             \end{subfigure}
             \hfill
             \begin{subfigure}{0.24\textwidth}
                 \centering
                 \vstretch{1.0}{\includegraphics[width=\textwidth]{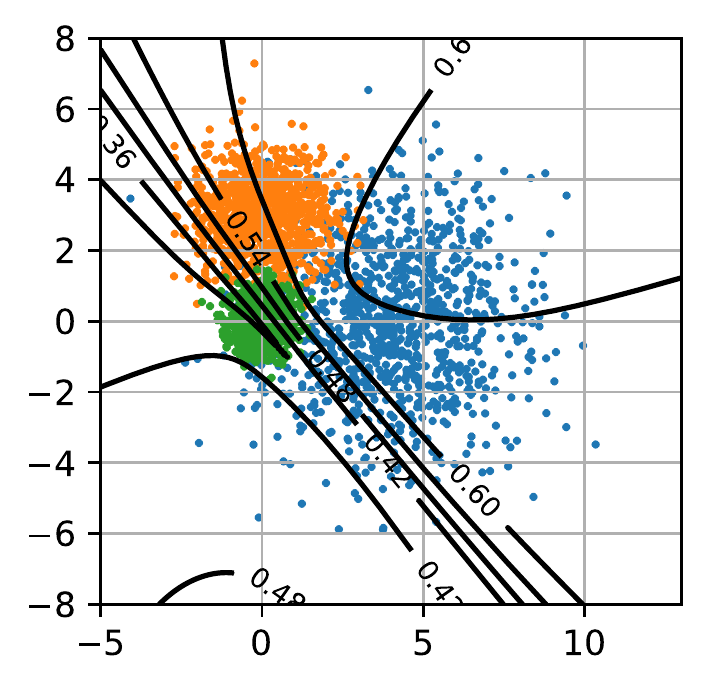}}
                 
                 \caption{Dirichlet conf.}
                 \label{fig:toy-dir-conf}
             \end{subfigure}
             \hfill
             \begin{subfigure}{0.24\textwidth}
                 \centering
                 \vstretch{1.0}{\includegraphics[width=\textwidth]{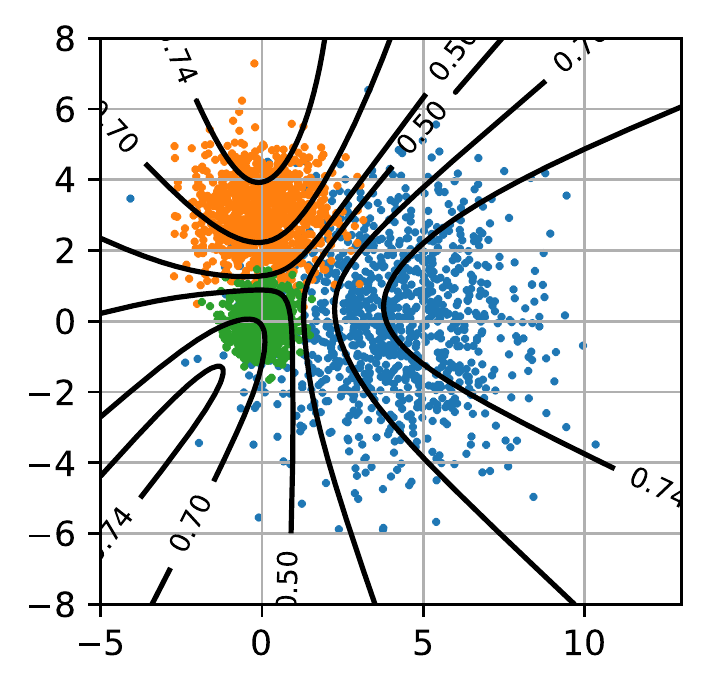}}
                 
                 \caption{Laplace conf.}
                 \label{fig:toy-lap-conf}
             \end{subfigure}

     \caption{An artificial three-class classification problem with 1000 examples per class. The top row shows the loss surface contours for various distillation approaches; darker colours imply lower losses. The bottom row shows the corresponding confidence contours (the confidence scores are reported on the contour). This shows that Dirichlet-based EDD is unable to learn properly whilst our proposed Laplace L-EDD can imitate an ensemble.}
     \label{fig:toy}
\end{figure*}

This section investigates the proposed Laplace logit-based ensemble distribution distillation (L-EDD) technique on a static artificial dataset, see Figure \ref{fig:toy-data}. The dataset was generated by sampling 3000 data points from three isotropic Gaussian distributions equally. The location and standard deviation of the Gaussians were chosen such that there would be regions with significant overlap and regions where models can be highly confident.

In these exploratory experiments, an ensemble of 10 small neural networks is first trained by randomly initialising each member. Thereafter, the ensemble is distilled using KD, EDD and Laplace L-EDD. We perform a qualitative comparison of these methods by displaying both the loss surface (see Figures \ref{fig:toy-dir-loss}-\ref{fig:toy-lap-loss}) of each approach and the resulting confidence (maximum softmax probability) contours ((see Figures \ref{fig:toy-ens-conf}-\ref{fig:toy-lap-conf})). 
The loss surface shows how the student distillation loss varies over the input space, thus providing useful information about which regions of the data are successfully optimised. 
The confidence contours are useful to understand if the system can separate between each of the three classes. EDD training had to be terminated early as it diverged due to large gradients originating from the high-confidence regions (as discussed in \citet{ryabin}).

Figure \ref{fig:toy-ens-conf} shows the ensemble confidence contours which clearly trace out class boundaries and partially separate the three classes. The confidence also increases as one moves further away from regions of overlap since there is less uncertainty. This is the behaviour we expect from a properly trained system and further, expect distilled students to behave similarly. Next, we knowledge distil the ensemble onto a single student model, see Figures \ref{fig:toy-dist-loss} and \ref{fig:toy-dist-conf}. The loss surface shows that the student can optimise the distillation objective over regions with high overlap well and generate confidence score contours that are consistent with the teacher ensemble. 

However, Ensemble Distribution Distillation completely fails on this very simple task. Observing the loss surface in Figure \ref{fig:toy-dir-loss}, one can infer that the Dirichlet student is unable to optimise regions where there is significant data overlap, instead displaying extremely small losses in regions for which the teacher ensemble is already confident. This links back to a result in \citet{ryabin} in which they find that highly confident teachers can induce extremely large gradients. This also translates into inaccurate confidence contours which are unable to separate between classes, especially in regions of overlap, see Figure \ref{fig:toy-dir-conf}. 

The Laplace L-EDD approach circumvents these issues by operating in the logit space. The resulting loss surface is much more consistent with the knowledge-distilled student, optimising regions of high overlap, see Figure \ref{fig:toy-lap-loss}. Similarly, the confidence contours trace out boundaries consistent with the ensemble but with lower overall confidence. Since the log-likelihood (similar to KL-divergence) is a mode-covering objective \citep{minka2005divergence}, the Laplace student ends up predicting distributions that overestimate the range of ensemble logits. Couple this with the long tails of the distributions and the Laplace will often overestimate the variance in logits and produce lower overall confidence.

\section{Machine Translation}
\label{sec:machine-translation}

This section reports on the performance of base transformers \citep{aiayn} trained on the En-De WMT'16 dataset, consisting of 4.5 million sentence pairs covering topics such as news \& policy.
We use newstest-13 for validation and newstest-14 for predictive evaluation.
For the main task of investigating out-of-distribution (OOD) detection, we compare the in-distribution (ID) newstest-14 with one of the publically available Khresmoi-Summary (Khresmoi) \citep{khresmoi}, MTNT \citep{mtnt} and Kyoto Free Translation Task (KFTT) \citep{kftt} datasets. These datasets relate to medical articles, Reddit-based noisy conversational text and specialised Wikipedia articles, respectively. 
Furthermore, we apply insights from training these systems to big transformers \citep{aiayn} trained on the larger En-Ru WMT'20 consisting of 58 million pairs after processing. In this case, we use newstest-19 for validation and newstest-20 for evaluation. The OOD detection task uses newstest-20 as ID and the same OOD datasets as above for the base transformer.

Data is tokenized using Moses, following \cite{scaling-nmt}. For WMT'16, a shared dictionary is trained using Byte Pair Encoding (BPE) with 32,000 merge operations \citep{bpe}. For WMT'20 we learn disjoint dictionaries using BPE with 40,000 merge operations.
The predictive performance (translation quality) will be evaluated using corpus-level (Sacre)BLEU \citep{sacrebleu}, with no post-processing of outputs before being scored. For the main task of detection, we use the ubiquitous threshold-independent AUROC metric \citep{auroc}, with baseline random detection corresponding to a score of 50\%.

All standard transformers are trained using an inverse square root with a linear warmup stage. A stronger Deep Ensemble baseline is formed by taking $M = 5$ such models. To avoid the high training cost of Deep Ensembles, we also train Snapshot Ensembles \citep{temporalensemble, snapshot} with a cyclic learning rate \citep{cyclic} to showcase that distribution distillation can be achievable with smaller training budgets. 
Since building Deep Ensembles is expensive, and the performance difference to Snapshot Ensembles was shown to be small, we opted to perform most distillation experiments on Snapshot Ensembles. We compare the proposed L-EDD approaches with KD and EDD and repeat each experiment 5 times, each with a different Snapshot Ensemble. 
Finally, similar to a range of prior work on uncertainty estimation tasks \citep{endd, structured, tcp}, we do not aim to achieve state-of-the-art predictive performance but opt for a simpler setup with a focus on achieving better uncertainty estimation. All setup details are provided in Appendix \ref{app:exp}. Hyperparameters were determined on ID validation sets.

\begin{table}[h!]
	\centering
	\begin{minipage}[t]{0.48\textwidth}%
		\begin{center}
                \caption{Model parameter size, relative training time and translation performance on newstest-14 $\pm$2std (BLEU) for base transformer. We include two different KD baselines, one for each teacher ensemble. 
                }
                \vspace{-2mm}
			\def\arraystretch{1.00}
			\makebox[1.0\textwidth][c]{
                    \small
				\begin{tabular}{l|cc|l}
					\toprule
					\multirow{2}{*}{\textbf{Model}} & 
					\multirow{2}{*}{\textbf{Size}} & 
					\textbf{Train} & 
					\multirow{2}{*}{\textbf{BLEU} $\uparrow$} \\ 
					& & \textbf{Time} & \\
					\midrule
					{{Standard}} & 60.9M & 1.0 & 25.85 \tpm 0.17 \\
					\midrule
					{{Deep Ensemble}} & 304.5M & 5.0 & 26.72 \\
					{KD (Categorical)} & 60.9M & 5.9 & 26.70 \tpm 0.26 \\
					\midrule
					{{Snapshot Ensemble}} & 304.5M  & 1.5 & 26.54 \tpm 0.16 \\
					{KD (Categorical)} & 60.9M & 1.9 & {27.02} \tpm 0.19 \\
					{EDD (Dirichlet)} & 60.9M & 2.0 & {26.96} \tpm 0.06 \\
					{L-EDD (Gaussian)} & 77.8M & 1.9 & {26.90} \tpm 0.28 \\
					{L-EDD (Laplace)} & 77.8M & 1.9 & {27.08} \tpm 0.20 \\
					\bottomrule
			\end{tabular}}
			\label{tab:wmt16}
		\end{center}
	\end{minipage}
\end{table}

\begin{table*}[b!]
	\centering{}
	\begin{minipage}[t]{1.0\textwidth}%
		\begin{center}
                \caption{OOD detection performance (\%{AUROC} $\uparrow$ $\pm$ 2 std) for base transformer with ID dataset newtest-14 and OOD datasets Khresmoi, MTNT and KFTT. \textbf{Bold} indicates best in a column, \underline{underline} second best. 
                Laplace L-EDD with knowledge uncertainty (KU), shows superior performance for all OOD datasets even compared to the Deep Ensemble.
                }
			\vspace{-2mm}
			\def\arraystretch{1.00}
			\small
			\begin{tabular}{l|ll|ll|ll}
				\toprule
				\multirow{2}{*}{\textbf{Model}} & 
				\multicolumn{2}{c|}{\textbf{Khresmoi}} & 
				\multicolumn{2}{c|}{\textbf{MTNT}} & 
				\multicolumn{2}{c}{\textbf{KFTT}}  \\
				& \multicolumn{1}{c}{\textbf{TU}} 
				& \multicolumn{1}{c|}{\textbf{KU}}  
				& \multicolumn{1}{c}{\textbf{TU}}  
				& \multicolumn{1}{c|}{\textbf{KU}}  
				& \multicolumn{1}{c}{\textbf{TU}}  
				& \multicolumn{1}{c}{\textbf{KU}}  \\
				\midrule
                    {Standard} & 47.5 \tpm 0.8 & \xmark & 63.5 \tpm 1.3 & \xmark & 30.6 \tpm 1.2 & \xmark \\
                    \midrule
				{Deep Ensemble} & 48.0 & 61.9 & 64.5 & 63.7 & 30.1 & 44.0 \\
                    {KD (Categorical)} & 47.9 \tpm 1.1 & \xmark & 64.5 \tpm 1.3 & \xmark & 29.8 \tpm 0.7 & \xmark \\
				\midrule
				Snapshot Ensemble  & 49.0 \tpm 0.6 & 62.6 \tpm 1.1 & 63.8 \tpm 1.2 & 63.1 \tpm 0.7 & 31.7 \tpm 0.9 & \underline{47.4} \tpm 2.5 \\
				{KD (Categorical)}  & 48.0 \tpm 1.4 & \xmark & 64.6 \tpm 0.9 & \xmark & 31.3 \tpm 0.5 & \xmark \\
				{EDD (Dirichlet)}   & 49.6 \tpm 1.3 & 57.1 \tpm 1.4 & \underline{65.1} \tpm 1.7 & \underline{65.6} \tpm 2.0 & 31.0 \tpm 0.9 & 36.2 \tpm 1.4 \\
				{L-EDD (Gaussian)}  & \underline{59.5} \tpm 1.1 & \underline{71.7} \tpm 1.9 & \textbf{66.3} \tpm 1.6 & 64.0 \tpm 2.1 & \underline{35.8} \tpm 1.2 & 44.0 \tpm 0.2 \\
				{L-EDD (Laplace)}   & \textbf{65.1} \tpm 1.8 & \textbf{73.1} \tpm 1.7 & \underline{65.1} \tpm 1.5 & \textbf{66.8} \tpm 1.8 & \textbf{37.8} \tpm 0.2 & \textbf{48.8} \tpm 1.4 \\
				\bottomrule
			\end{tabular}
			\label{tab:wmt16-detection}
		\end{center}
	\end{minipage}
\end{table*} 

\subsection{Base Transformer Results}
\label{sec:res-base}

Table \ref{tab:wmt16} shows both the efficiency and performance of a wide range of systems on newstest-14. As expected the performance of the Deep Ensemble trumps both the Snapshot Ensemble and a standard trained system. Surprisingly, Snapshot Ensemble distilled students achieve better performance, a pattern also observed in self-distilled systems and is explored in more detail in \citet{selfmicro}. 

Next, we compare the threshold-independent out-of-distribution detection performance of baseline systems with L-EDD models.
From Table \ref{tab:wmt16-detection}, we observe that Snapshot Ensembles are able to compete with the Deep equivalent whilst being more than 3 times cheaper to train.
Furthermore, the knowledge-distilled students are able to match the detection performance of their Deep and Snapshot ensemble teachers using total uncertainty (TU). This is a natural result since they were specifically designed to capture the predictive distribution of their teacher ensemble. However, since KD students are unable to estimate knowledge uncertainty (KU), they fail to reach ensemble-level detection performance in all but the MTNT dataset.
Similarly, the modified Dirichlet baseline as described by \citet{ryabin} is able to achieve similar detection performance using TU but with the added ability to estimate KU. And whilst the Dirichlet KU are often better than its TU estimates, they often fall short when compared to ensembles.

On the other hand, the Laplace \& Gaussian L-EDD models are (surprisingly) able to outperform both ensembles in all three detection splits, producing either similar or significantly better TU and KU estimates. This may partially be due to the fact that diagonal Laplace and Gaussian distributions have more parameters and are more flexible than the Dirichlet, and also because they do not suffer from the same optimisation issues.
Nonetheless, neither reason explains why L-EDD models can outperform ensembles in detection. We explore this pattern in Section \ref{sec:analysis}.
Additionally, many models are worse than a random detector, especially for the KFTT and partially for the Khresmoi dataset. A partial explanation could be that these datasets contain longer sequences. When decoding, the transformer models produce more and more confident predictions further along in the output sequence, causing lower uncertainty scores. Section \ref{sec:analysis} investigates this effect and isolates a possible reason behind Laplace's success in outperforming ensembles.

As an aside, we observe that estimates of knowledge uncertainty are clearly important for OOD detection for autoregressive sequence tasks (and this is corroborated in prior work \cite{structured, ryabin}). This is in contrast to recent empirical results on image classification data, which show the opposite, that measures of knowledge uncertainty are not useful for indicating distributional shifts \citep{Xia2022OnTU, abe2022deep}. 

\begin{table}[h!]
	\centering
	\begin{minipage}[t]{0.48\textwidth}%
		\begin{center}
                \caption{Model parameter size, relative training time and translation performance on newstest-20 $\pm$2std (BLEU) for the big transformer. 
                }
                \vspace{-2mm}
			\def\arraystretch{1.00}
			\makebox[1.0\textwidth][c]{
                    \small
				\begin{tabular}{l|cc|l}
					\toprule
					\multirow{2}{*}{\textbf{Model}} & 
					\multirow{2}{*}{\textbf{Size}} & 
					\textbf{Train} & 
					\multirow{2}{*}{\textbf{BLEU} $\uparrow$} \\ 
					& & \textbf{Time} & \\
					\midrule
					{{Standard}} & 271M & 1.0 & 26.28 \tpm 0.34 \\
					\midrule
					{{Deep Ensemble}} & 1.35B & 5.0 & 26.81 \\
					\midrule
					{{Snapshot Ensemble}} & 1.35B  & 1.5 & 26.42 \tpm 0.23 \\
                    {KD (Categorical)} & 271M & 2.1 & 26.73 \tpm 0.16 \\
					{EDD (Dirichlet)} & 271M & 2.2 & 26.66 \tpm 0.19 \\
                    {L-EDD (Laplace)} & 320M & 2.2 & 26.71 \tpm 0.18 \\
					\bottomrule
			\end{tabular}}
			\label{tab:wmt20}
		\end{center}
	\end{minipage}
\end{table}

\begin{table*}[b!]
	\centering{}
	\begin{minipage}[t]{1.0\textwidth}%
		\begin{center}
			\caption{OOD detection performance (\%{AUROC} $\uparrow$ $\pm$ 2 std) for big transformer with ID dataset newtest-20 and OOD datasets Khresmoi, MTNT and KFTT. \textbf{Bold} indicates best in a column, \underline{underline} second best. Similar to Table \ref{tab:wmt20-detection}, L-EDD (Laplace) with KU shows superior performance over all OOD datasets.}
			\vspace{-2mm}
			\def\arraystretch{1.00}
			\small
			\begin{tabular}{l|ll|ll|ll}
				\toprule
				\multirow{2}{*}{\textbf{Model}} & 
				\multicolumn{2}{c|}{\textbf{Khresmoi}} & 
				\multicolumn{2}{c|}{\textbf{MTNT}} & 
				\multicolumn{2}{c}{\textbf{KFTT}}  \\
				& \multicolumn{1}{c}{\textbf{TU}} 
				& \multicolumn{1}{c|}{\textbf{KU}}  
				& \multicolumn{1}{c}{\textbf{TU}}  
				& \multicolumn{1}{c|}{\textbf{KU}}  
				& \multicolumn{1}{c}{\textbf{TU}}  
				& \multicolumn{1}{c}{\textbf{KU}}  \\
				\midrule
				{Deep Ensemble} & 39.3 & 53.2 & 70.8 & 69.0 & 51.0 & 60.3 \\
				\midrule
                    Snapshot Ensemble  & \underline{40.8} \tpm 0.5 & \underline{55.0} \tpm 0.8 & 70.1 \tpm 0.5 & \underline{69.3} \tpm 0.9 & \underline{51.1} \tpm 0.6 & \underline{60.9} \tpm 1.4 \\
                    {KD (Categorical)}		   & 40.4 \tpm 0.8 & \xmark & \underline{70.9} \tpm 1.0 & \xmark & 50.9 \tpm 0.6 & \xmark \\
                    {EDD (Dirichlet)} & 41.0 \tpm 0.8 & 52.9 \tpm 1.3 & 71.3 \tpm 0.7 & 69.4 \tpm 0.7 & 51.0 \tpm 0.8 & 60.0 \tpm 1.3 \\
                    {L-EDD (Laplace)}	   & \textbf{51.0} \tpm 0.9 & \textbf{63.4} \tpm 1.2 & \textbf{72.6} \tpm 0.8 & \textbf{70.2} \tpm 0.6 & \textbf{63.2} \tpm 1.0 & \textbf{70.2} \tpm 1.1 \\
				\bottomrule
			\end{tabular}
			\label{tab:wmt20-detection}
		\end{center}
	\end{minipage}
\end{table*} 

\subsection{Big Transformer Results}
\label{sec:res-big}

In this section, we take the best-performing systems from the previous section and apply them to the `big transformer' on the larger En-Ru WMT'20 dataset. Table \ref{tab:wmt20} shows the efficiency and predictive performance on newstest-20. Again, we observe that the Deep Ensemble outperforms its Snapshot equivalent. Furthermore, the KD and L-EDD students, distilled from the Snapshot Ensemble were able to outperform their teacher. However, unlike the smaller-scale experiments, these students were only able to reach Deep Ensemble performance within a standard deviation, but were able to do so with a single forward pass.

From Table \ref{tab:wmt20-detection} we observe a similar pattern in which Deep and Snapshot ensembles perform equivalently whilst L-EDD (Laplace) is able to significantly outperform both ensembles in all but the MTNT dataset. Interestingly, unlike in Table \ref{tab:wmt16-detection} where no model was able to beat a random detector on the KFTT detection, the larger En-Ru WMT'20 based models are able to differentiate between newstest-20 and KFTT; switching the ID dataset to newstest-14 does not affect the results notably.

\section{Analysis: Ensemble vs Laplace}
\label{sec:analysis}

\subsection{Augmented Ensemble Uncertainties}

Both Sections \ref{sec:res-base} and \ref{sec:res-big} found that L-EDD models overall significantly outperformed their teacher Snapshot Ensemble and a Deep Ensemble. Therefore, we propose an alternative experiment to understand the source of L-EDD's superior performance. We fit an auxiliary Laplace distribution to a Deep Ensemble during inference and use the samples from this proxy to perform the detection task.

Consider a Deep Ensemble which produces a set of normalised logits $\{\tilde{\bm z}_{l}^{(m)}\}_{m=1}^M$ as in Equation (\ref{eq:logits}). In traditional uncertainty estimation, these logits would be transformed into categorical distributions. However, in this experiment, we estimate an auxiliary Laplace distribution using maximum likelihood (which is the loss-minimising distribution for a Laplace L-EDD student):
\begin{align}
    \tilde{\bm\mu}_l, \tilde{\bm\sigma}_l = \argmax_{\bm\mu, \bm\sigma} \sum_{m} \ln{\tt Lap}(\tilde{\bm z}_l^{(m)} \vert \bm\mu, \bm\sigma).
\end{align}
By sampling new points from this auxiliary distribution, we can estimate total and knowledge uncertainty:
\begin{align}
    \tilde{\bm\pi} = \Softmax(\bm z), \thickspace \bm z \sim {\tt Lap}(\tilde{\bm\mu}_l, \tilde{\bm\sigma}_l).
\end{align}
The aim of this modified approach to ensemble-based uncertainty estimation is to investigate whether or not approximating the logits with a Laplace distribution is the reason behind L-EDD performing better. 

\begin{table*}[t!]
	\centering{}
	\begin{minipage}[t]{1.0\textwidth}%
		\begin{center}
			\caption{OOD detection performance (\%{AUROC} $\uparrow$ $\pm$ 2 std) following the same setup as in Table \ref{tab:wmt16-detection}. The Laplace augmented ensemble demonstrates much better performance in most cases compared to its standard counterpart.}
			\vspace{-2mm}
			\def\arraystretch{1.00}
			\small
			\begin{tabular}{l|ll|ll|ll}
				\toprule
				\multirow{2}{*}{\textbf{Model}} & 
				\multicolumn{2}{c|}{\textbf{Khresmoi}} & 
				\multicolumn{2}{c|}{\textbf{MTNT}} & 
				\multicolumn{2}{c}{\textbf{KFTT}}  \\
				& \multicolumn{1}{c}{\textbf{TU}} 
				& \multicolumn{1}{c|}{\textbf{KU}}  
				& \multicolumn{1}{c}{\textbf{TU}}  
				& \multicolumn{1}{c|}{\textbf{KU}}  
				& \multicolumn{1}{c}{\textbf{TU}}  
				& \multicolumn{1}{c}{\textbf{KU}}  \\
				\midrule
                    {Deep Ensemble}           & 48.0 & 61.9 & \underline{64.5} & \underline{63.7} & 30.1 & 44.0 \\
                    {Deep Ensemble (Laplace)} & \underline{62.5} & \underline{72.1} & 63.8 & 56.1 & \underline{34.8} & \textbf{55.4} \\
				\midrule
                    {L-EDD (Laplace)}         & \textbf{65.1} \tpm 1.8 & \textbf{73.1} \tpm 1.7 & \textbf{65.1} \tpm 1.5 & \textbf{66.8} \tpm 1.8 & \textbf{37.8} \tpm 0.2 & \underline{48.8} \tpm 1.4 \\
				\bottomrule
			\end{tabular}
			\label{tab:wmt16-detection-laplace}
		\end{center}
	\end{minipage}
\end{table*} 

Table \ref{tab:wmt16-detection-laplace} shows the detection performance of the Laplace-modified Deep Ensemble, following the detection setup in Section \ref{sec:res-base}. 
Clearly, the augmented ensemble bridges the OOD detection performance gap between standard Deep Ensemble and L-EDD.
This suggests that measures of uncertainty based on directly fit, logit-space models of an ensemble are better at indicating distributional shift \textit{than directly using the ensemble logits}, for autoregressive sequence prediction. We remark that fully understanding why this is the case would be an interesting direction of future research, as it may enable further advancement in autoregressive out-of-distribution detection.
%


\subsection{Model Confidence for Longer Sequences}

Another reason for Laplace L-EDD's superior performance could be found in analysing behaviour with increasing sequence lengths.
\begin{figure*}[b!]
     \centering
     \begin{subfigure}[b]{0.24\textwidth}
         \centering
         \includegraphics[width=\textwidth]{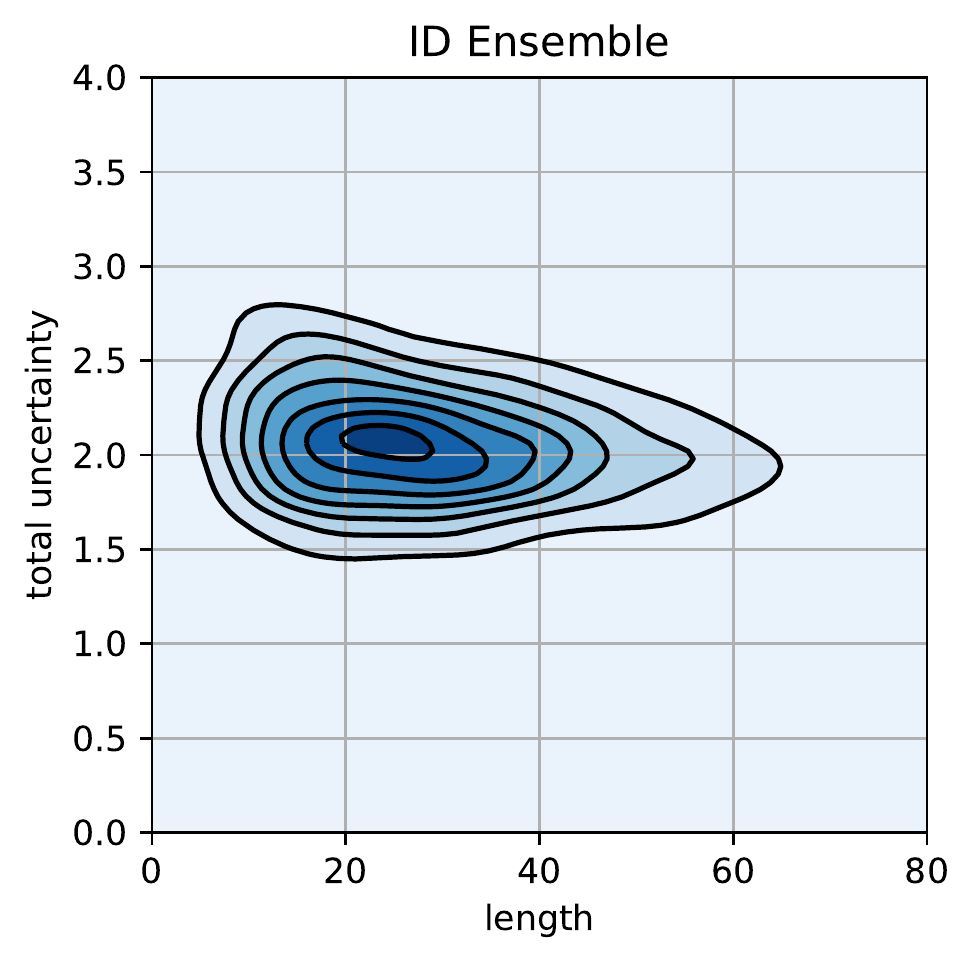}
         \vspace*{-6mm}
         \caption{PCC = -20\%}
         \vspace*{2mm}
     \end{subfigure}
     \hfill
     \begin{subfigure}[b]{0.24\textwidth}
         \centering
         \includegraphics[width=\textwidth]{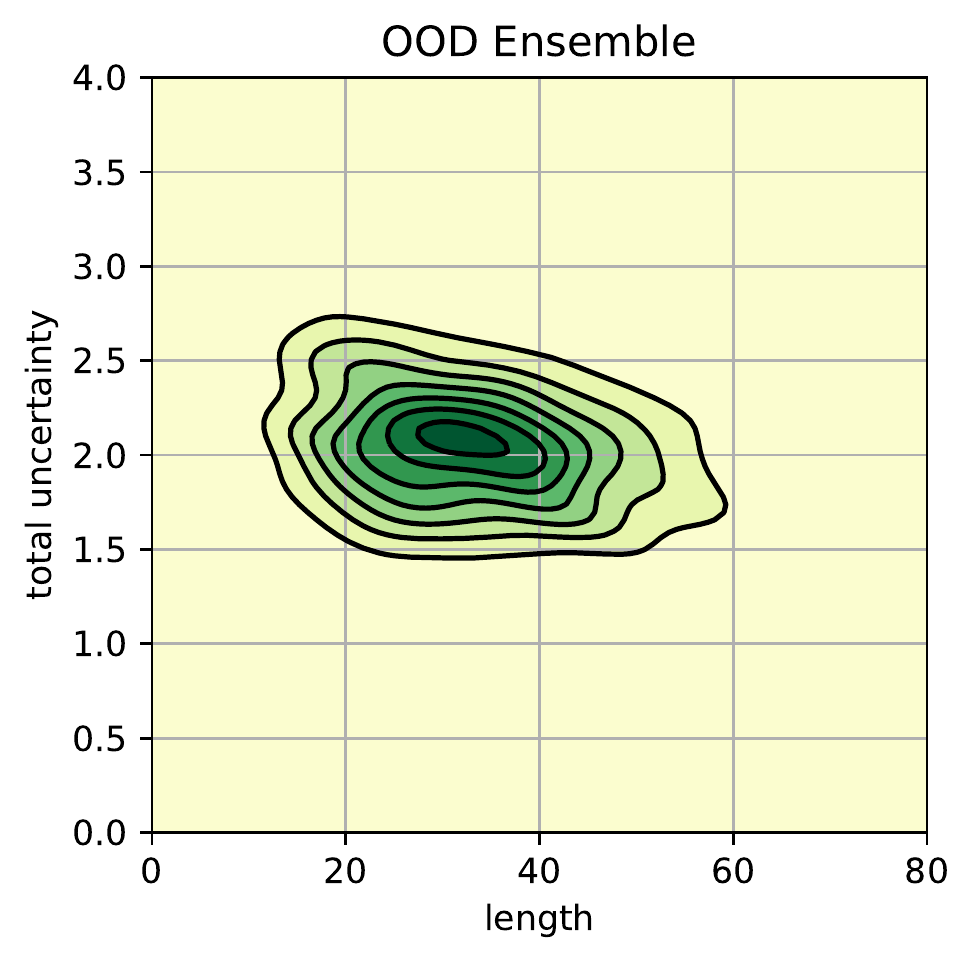}
         \vspace*{-6mm}
         \caption{PCC = -29\%}
         \vspace*{2mm}
     \end{subfigure}
     \hfill
     \begin{subfigure}[b]{0.24\textwidth}
         \centering
         \includegraphics[width=\textwidth]{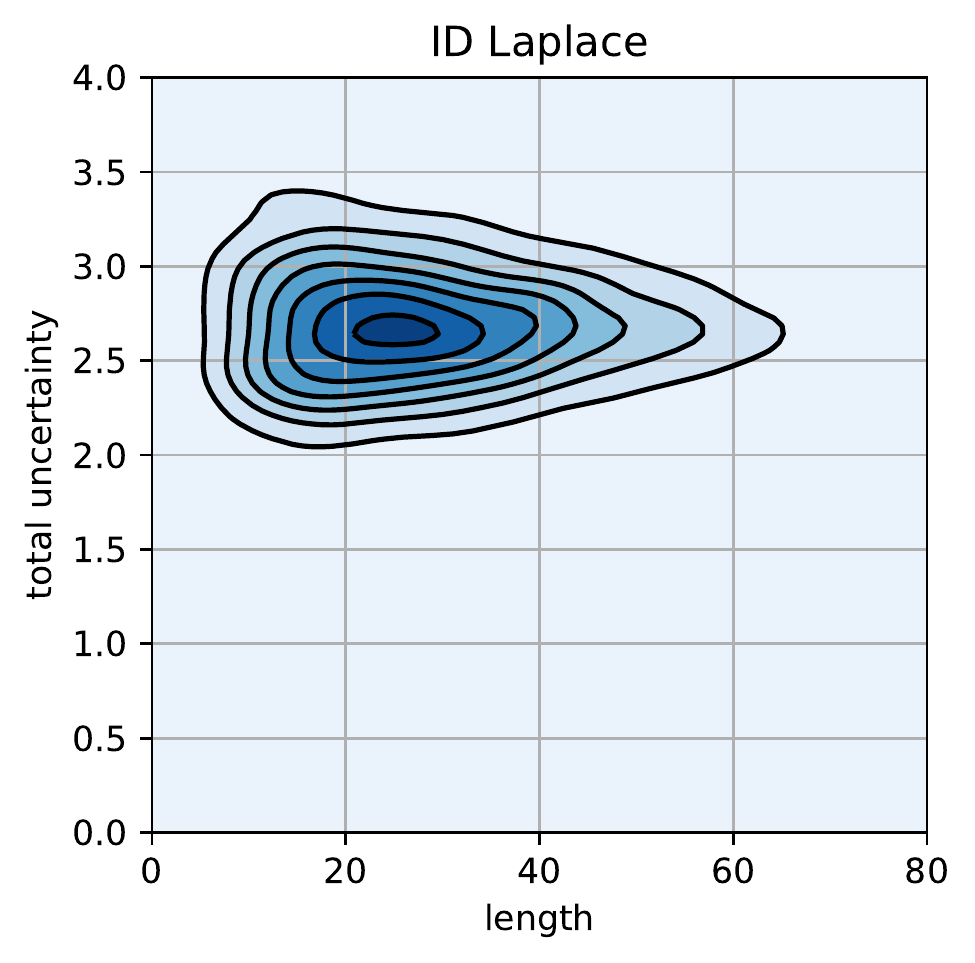}
         \vspace*{-6mm}
         \caption{PCC = -3\%}
         \vspace*{2mm}
     \end{subfigure}
     \hfill
     \begin{subfigure}[b]{0.24\textwidth}
         \centering
         \includegraphics[width=\textwidth]{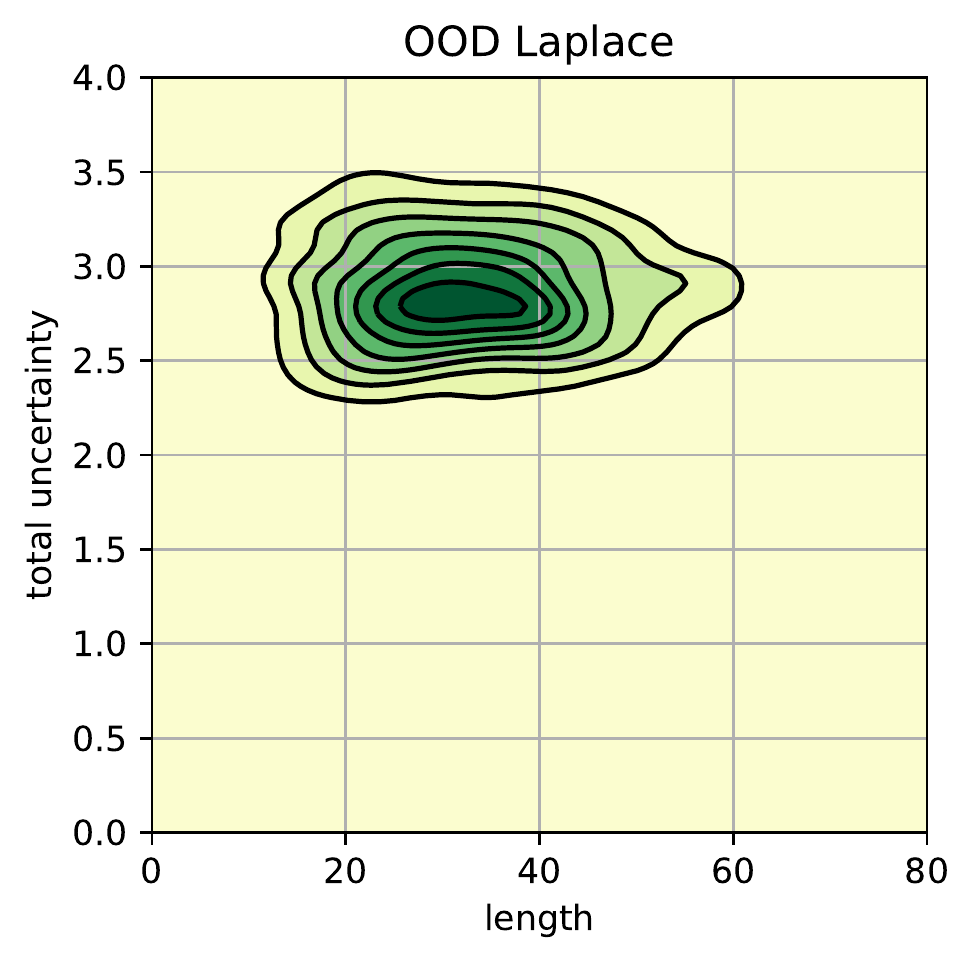}
         \vspace*{-6mm}
         \caption{PCC = 1\%}
         \vspace*{2mm}
     \end{subfigure}
     \vspace*{-2mm}
     \caption{Length vs uncertainty density plots for ID (newstest-14) and OOD (Khresmoi) datasets. The left half corresponds to the Deep Ensemble and the right half to Laplace L-EDD. Each figure caption also shows the Pearson Correlation Coefficient (PCC) between sequence length and total uncertainty. Total uncertainty and length are negatively correlated for the ensemble, i.e. it is more confident on longer sequences, but they are uncorrelated for L-EDD.}
     \label{fig:confirmation-bias}
\end{figure*}
Figure \ref{fig:confirmation-bias} shows how Deep Ensemble and L-EDD total uncertainties scale with output sequence length for both ID (newstest-14) and OOD (Khresmoi) datasets. Under each figure is also the associated Pearson correlation.

Three observations can be made from this data. The first is that the L-EDD system consistently outputs much higher total uncertainties. The second is that the Deep Ensemble displays a negative correlation between total uncertainty and sequence length. This implies that the ensemble is more confident in translating longer sequences, but this is also why it fails in detecting OOD datasets which contain longer sequences.
The third and possibly most significant observation is that the Laplace system shows almost no correlation between total uncertainty and sequence length for both ID and OOD datasets, effectively eliminating the length bias, and allowing it to better differentiate between ID and OOD datasets even when the OOD inputs differ in length from what the detection system was trained on.

\section{Conclusion}
\label{sec:conclusion}
In this work, we investigate the efficient estimation of uncertainties for large-scale autoregressive sequence prediction. To this end, we examine Ensemble Distribution Distillation (EDD) in the \textit{logit}-space, in order to bypass optimisation issues found in softmax-space EDD.
We perform experiments using modern transformer models trained to perform large-scale machine translation. They show that a student model trained to parameterise a \textit{Laplace} distribution over logits is able to significantly outperform Deep Ensembles for OOD detection at a fraction of the inference cost, whilst matching the ensemble for translation quality. 
Moreover, we show that the use of Snapshot Ensembling can greatly reduce the training costs of EDD, without sacrificing translation performance.
We hope that our work can encourage further investigation into the comparatively less well-explored domain of uncertainty estimation for structured sequence prediction, on tasks such as machine translation, image captioning, and automatic speech recognition. 

\begin{acknowledgements} 
    
Guoxuan Xia is funded jointly by Arm ltd. and EPSRC.
\end{acknowledgements}
\bibliography{bibliography}

\begin{thebibliography}{48}
\providecommand{\natexlab}[1]{#1}
\providecommand{\url}[1]{\texttt{#1}}
\expandafter\ifx\csname urlstyle\endcsname\relax
  \providecommand{\doi}[1]{doi: #1}\else
  \providecommand{\doi}{doi: \begingroup \urlstyle{rm}\Url}\fi

\bibitem[Abe et~al.(2022)Abe, Buchanan, Pleiss, Zemel, and
  Cunningham]{abe2022deep}
Taiga Abe, E.~Kelly Buchanan, Geoff Pleiss, Richard Zemel, and John~Patrick
  Cunningham.
\newblock Deep ensembles work, but are they necessary?
\newblock In \emph{Advances in Neural Information Processing Systems}, 2022.

\bibitem[Allen-Zhu and Li(2021)]{selfmicro}
Zeyuan Allen-Zhu and Yuanzhi Li.
\newblock Towards understanding ensemble, knowledge distillation and
  self-distillation in deep learning.
\newblock In \emph{arXiv}, 2021.

\bibitem[Bahdanau et~al.(2015)Bahdanau, Cho, and Bengio]{atten}
Dzmitry Bahdanau, Kyunghyun Cho, and Yoshua Bengio.
\newblock Neural machine translation by jointly learning to align and
  translate.
\newblock In \emph{International Conference on Learning Representations}, 2015.

\bibitem[Corbi\`{e}re et~al.(2019)Corbi\`{e}re, Thome, Bar-Hen, Cord, and
  P\'{e}rez]{tcp}
Charles Corbi\`{e}re, Nicolas Thome, Avner Bar-Hen, Matthieu Cord, and Patrick
  P\'{e}rez.
\newblock Addressing failure prediction by learning model confidence.
\newblock In \emph{Advances in Neural Information Processing Systems 32}. 2019.

\bibitem[Depeweg et~al.(2018)Depeweg, Hernández-Lobato, Doshi-Velez, and
  Udluft]{decomp}
Stefan Depeweg, Jose~Miguel Hernández-Lobato, Finale Doshi-Velez, and Steffen
  Udluft.
\newblock Decomposition of uncertainty in bayesian deep learning for efficient
  and risk-sensitive learning.
\newblock In \emph{International Conference on Learning Representations}, 2018.

\bibitem[Du{\v s}ek et~al.(2017)Du{\v s}ek, Haji{\v c}, Hlav{\'a}{\v c}ov{\'a},
  Libovick{\'y}, Pecina, Tamchyna, and Ure{\v s}ov{\'a}]{khresmoi}
Ond{\v r}ej Du{\v s}ek, Jan Haji{\v c}, Jaroslava Hlav{\'a}{\v c}ov{\'a},
  Jind{\v r}ich Libovick{\'y}, Pavel Pecina, Ale{\v s} Tamchyna, and Zde{\v
  n}ka Ure{\v s}ov{\'a}.
\newblock Khresmoi summary translation test data 2.0.
\newblock \url{http://hdl.handle.net/11234/1-2122}, 2017.
\newblock {LINDAT}/{CLARIAH}-{CZ} digital library at the Institute of Formal
  and Applied Linguistics ({{\'U}FAL}).

\bibitem[Fathullah and Gales(2022)]{s2d}
Yassir Fathullah and Mark J.~F. Gales.
\newblock Self-distribution distillation: efficient uncertainty estimation.
\newblock In \emph{Proceedings of the Thirty-Eighth Conference on Uncertainty
  in Artificial Intelligence}, volume 180 of \emph{Proceedings of Machine
  Learning Research}, 01--05 Aug 2022.

\bibitem[Fathullah et~al.(2021)Fathullah, Gales, and Malinin]{seq-endd}
Yassir Fathullah, Mark J.~F. Gales, and Andrey Malinin.
\newblock Ensemble distillation approaches for grammatical error correction.
\newblock In \emph{International Conference on Acoustics, Speech, and Signal
  Processing}, 2021.

\bibitem[Gaido et~al.(2020)Gaido, Gangi, Negri, and Turchi]{seq-kd-app1}
Marco Gaido, Mattia A.~Di Gangi, Matteo Negri, and Marco Turchi.
\newblock End-to-end speech-translation with knowledge distillation:
  {FBK}@{IWSLT}2020.
\newblock In \emph{International Conference on Spoken Language Translation},
  2020.

\bibitem[Gal et~al.(2017)Gal, Islam, and Ghahramani]{pmlr-v70-gal17a}
Yarin Gal, Riashat Islam, and Zoubin Ghahramani.
\newblock Deep {B}ayesian active learning with image data.
\newblock In \emph{Proceedings of the 34th International Conference on Machine
  Learning}, volume~70 of \emph{Proceedings of Machine Learning Research},
  06--11 Aug 2017.

\bibitem[Geifman and El-Yaniv(2019)]{pmlr-v97-geifman19a}
Yonatan Geifman and Ran El-Yaniv.
\newblock {S}elective{N}et: A deep neural network with an integrated reject
  option.
\newblock In \emph{Proceedings of the 36th International Conference on Machine
  Learning}, volume~97 of \emph{Proceedings of Machine Learning Research},
  09--15 Jun 2019.

\bibitem[Gustafsson et~al.(2020)Gustafsson, Danelljan, and
  Sch{\"o}n]{gustafsson2020evaluating}
Fredrik~K Gustafsson, Martin Danelljan, and Thomas~B Sch{\"o}n.
\newblock Evaluating scalable bayesian deep learning methods for robust
  computer vision.
\newblock In \emph{Proceedings of the IEEE/CVF Conference on Computer Vision
  and Pattern Recognition (CVPR) Workshops}, 2020.

\bibitem[Hinton et~al.(2014)Hinton, Vinyals, and Dean]{kd}
Geoffrey Hinton, Oriol Vinyals, and Jeff Dean.
\newblock Distilling the knowledge in a neural network.
\newblock In \emph{Conference on Neural Information Processing Systems Deep
  Learning Workshop}, 2014.

\bibitem[Huang et~al.(2017)Huang, Li, Pleiss, Liu, Hopcroft, and
  Weinberger]{snapshot}
Gao Huang, Yixuan Li, Geoff Pleiss, Zhuang Liu, John~E. Hopcroft, and Kilian~Q.
  Weinberger.
\newblock Snapshot ensembles: Train 1, get m for free.
\newblock In \emph{International Conference on Learning Representations}, 2017.

\bibitem[Hullermeier and Waegeman(2021)]{duku}
Eyke Hullermeier and Willem Waegeman.
\newblock Aleatoric and epistemic uncertainty in machine learning: An
  introduction to concepts and methods.
\newblock In \emph{Machine Learning}, 2021.

\bibitem[Jiao et~al.(2019)Jiao, Yin, Shang, Jiang, Chen, Li, Wang, and
  Liu]{seq-kd-app3}
Xiaoqi Jiao, Yichun Yin, Lifeng Shang, Xin Jiang, Xiao Chen, Linlin Li, Fang
  Wang, and Qun Liu.
\newblock {T}iny{BERT}: Distilling {BERT} for natural language understanding.
\newblock In \emph{International Conference on Learning Representations}, 2019.

\bibitem[Kim et~al.(2021)Kim, Koo, and Hwang]{Kim2021AUB}
Jihyo Kim, Jiin Koo, and Sangheum Hwang.
\newblock A unified benchmark for the unknown detection capability of deep
  neural networks.
\newblock \emph{ArXiv}, abs/2112.00337, 2021.

\bibitem[Kim and Rush(2016)]{seq-kd}
Yoon Kim and Alexander~M. Rush.
\newblock Sequence-level knowledge distillation.
\newblock In \emph{Conference on Empirical Methods in Natural Language
  Processing}, 2016.

\bibitem[Kingma and Ba(2015)]{adam}
Diederik~P. Kingma and Jimmy Ba.
\newblock Adam: A method for stochastic optimization.
\newblock In \emph{International Conference for Learning Representations},
  2015.

\bibitem[Lakshminarayanan et~al.(2017)Lakshminarayanan, Pritzel, and
  Blundell]{deepens}
Balaji Lakshminarayanan, Alexander Pritzel, and Charles Blundell.
\newblock Simple and scalable predictive uncertainty estimation using deep
  ensembles.
\newblock In \emph{Conference on Neural Information Processing Systems}, 2017.

\bibitem[Lee et~al.(2018)Lee, Lee, Lee, and Shin]{ood-training-1}
Kimin Lee, Honglak Lee, Kibok Lee, and Jinwoo Shin.
\newblock Training confidence-calibrated classifiers for detecting
  out-of-distribution samples.
\newblock In \emph{International Conference on Learning Representations}, 2018.

\bibitem[Lindqvist et~al.(2020)Lindqvist, Olmin, Lindsten, and
  Svensson]{mod-edd}
Jakob Lindqvist, Amanda Olmin, Fredrik Lindsten, and Lennart Svensson.
\newblock A general framework for ensemble distribution distillation.
\newblock In \emph{International Workshop on Machine Learning for Signal
  Processing (MLSP)}, 2020.

\bibitem[Liu et~al.(2020)Liu, Lin, Padhy, Tran, Bedrax~Weiss, and
  Lakshminarayanan]{sngp}
Jeremiah Liu, Zi~Lin, Shreyas Padhy, Dustin Tran, Tania Bedrax~Weiss, and
  Balaji Lakshminarayanan.
\newblock Simple and principled uncertainty estimation with deterministic deep
  learning via distance awareness.
\newblock In \emph{Advances in Neural Information Processing Systems},
  volume~33, 2020.

\bibitem[Malinin and Gales(2021)]{structured}
Andrey Malinin and Mark J.~F. Gales.
\newblock Uncertainty estimation in autoregressive structured prediction.
\newblock In \emph{International Conference on Learning Representations}, 2021.

\bibitem[Malinin et~al.(2017)Malinin, Ragni, Knill, and Gales]{ood-training-2}
Andrey Malinin, Anton Ragni, Kate Knill, and Mark J.~F. Gales.
\newblock Incorporating uncertainty into deep learning for spoken language
  assessment.
\newblock In \emph{Association for Computational Linguistics}, 2017.

\bibitem[Malinin et~al.(2020)Malinin, Mlodozeniec, and Gales]{endd}
Andrey Malinin, Bruno Mlodozeniec, and Mark J.~F. Gales.
\newblock Ensemble distribution distillation.
\newblock In \emph{International Conference on Learning Representations}, 2020.

\bibitem[Manning and Schütze(1999)]{auroc}
Chris Manning and Hinrich Schütze.
\newblock \emph{Foundations of Statistical Natural Language Processing}.
\newblock MIT Press, 1999.

\bibitem[Michel and Neubig(2018)]{mtnt}
Paul Michel and Graham Neubig.
\newblock Mtnt: A testbed for machine translation of noisy text.
\newblock In \emph{Conference on Empirical Methods in Natural Language
  Processing}, 2018.

\bibitem[Minka et~al.(2005)]{minka2005divergence}
Tom Minka et~al.
\newblock Divergence measures and message passing.
\newblock Technical report, Technical report, Microsoft Research, 2005.

\bibitem[Moon et~al.(2020)Moon, Kim, Shin, and
  Hwang]{Moon2020ConfidenceAwareLF}
Jooyoung Moon, Jihyo Kim, Younghak Shin, and Sangheum Hwang.
\newblock Confidence-aware learning for deep neural networks.
\newblock In \emph{International Conference on Machine Learning}, 2020.

\bibitem[Neubig(2011)]{kftt}
Graham Neubig.
\newblock The {Kyoto} free translation task.
\newblock \url{http://www.phontron.com/kftt}, 2011.

\bibitem[Notin et~al.(2021)Notin, Hern{\'a}ndez-Lobato, and
  Gal]{notin2021improving}
Pascal Notin, Jos{\'e}~Miguel Hern{\'a}ndez-Lobato, and Yarin Gal.
\newblock Improving black-box optimization in {VAE} latent space using decoder
  uncertainty.
\newblock In \emph{Advances in Neural Information Processing Systems}, 2021.

\bibitem[Ott et~al.(2018)Ott, Edunov, Grangier, and Auli]{scaling-nmt}
Myle Ott, Sergey Edunov, David Grangier, and Michael Auli.
\newblock Scaling neural machine translation.
\newblock In \emph{Proceedings of the Third Conference on Machine Translation:
  Research Papers}, 2018.

\bibitem[Ott et~al.(2019)Ott, Edunov, Baevski, Fan, Gross, Ng, Grangier, and
  Auli]{fairseq}
Myle Ott, Sergey Edunov, Alexei Baevski, Angela Fan, Sam Gross, Nathan Ng,
  David Grangier, and Michael Auli.
\newblock fairseq: A fast, extensible toolkit for sequence modeling.
\newblock In \emph{Proceedings of NAACL-HLT 2019: Demonstrations}, 2019.

\bibitem[Ovadia et~al.(2019)Ovadia, Fertig, Ren, Nado, Sculley, Nowozin,
  Dillon, Lakshminarayanan, and Snoek]{ovadia_trust}
Yaniv Ovadia, Emily Fertig, Jie Ren, Zachary Nado, D.~Sculley, Sebastian
  Nowozin, Joshua Dillon, Balaji Lakshminarayanan, and Jasper Snoek.
\newblock Can you trust your model\textquotesingle s uncertainty? evaluating
  predictive uncertainty under dataset shift.
\newblock In \emph{Advances in Neural Information Processing Systems},
  volume~32, 2019.

\bibitem[Post(2018)]{sacrebleu}
Matt Post.
\newblock A call for clarity in reporting {BLEU} scores.
\newblock In \emph{Association for Computational Linguistics}, 2018.

\bibitem[Radmard et~al.(2021)Radmard, Fathullah, and Lipani]{seq-al2}
Puria Radmard, Yassir Fathullah, and Aldo Lipani.
\newblock Subsequence based deep active learning for named entity recognition.
\newblock In \emph{Association for Computational Linguistics}, 2021.

\bibitem[Ryabinin et~al.(2021)Ryabinin, Malinin, and Gales]{ryabin}
Max Ryabinin, Andrey Malinin, and Mark J.~F. Gales.
\newblock Scaling ensemble distribution distillation to many classes with proxy
  targets.
\newblock In \emph{Conference on Neural Information Processing Systems}, 2021.

\bibitem[Sennrich et~al.(2016)Sennrich, Haddow, and Birch]{bpe}
Rico Sennrich, Barry Haddow, and Alexandra Birch.
\newblock Neural machine translation of rare words with subword units.
\newblock In \emph{Association for Computational Linguistics}, 2016.

\bibitem[Smith(2017)]{cyclic}
Leslie~N. Smith.
\newblock Cyclical learning rates for training neural networks.
\newblock In \emph{Winter Conference on Applications of Computer Vision}, 2017.

\bibitem[Tan et~al.(2019)Tan, Ren, He, Qin, and Liu]{seq-kd-app2}
Xu~Tan, Yi~Ren, Di~He, Tao Qin, and Tie-Yan Liu.
\newblock Multilingual neural machine translation with knowledge distillation.
\newblock In \emph{International Conference on Learning Representations}, 2019.

\bibitem[Vaswani et~al.(2017)Vaswani, Shazeer, Parmar, Uszkoreit, Jones, Gomez,
  Kaiser, and Polosukhin]{aiayn}
Ashish Vaswani, Noam Shazeer, Niki Parmar, Jakob Uszkoreit, Llion Jones,
  Aidan~N. Gomez, Lukasz Kaiser, and Illia Polosukhin.
\newblock Attention is all you need.
\newblock In \emph{Conference on Neural Information Processing Systems}, 2017.

\bibitem[Wong et~al.(2016)Wong, Gales, and Wang]{gen-student-th}
Jeremy H.~M. Wong, Mark J.~F. Gales, and Yu~Wang.
\newblock Sequence-level knowledge distillation.
\newblock In \emph{IEEE/ACM Transactions on Audio Speech and Language
  Processing}, 2016.

\bibitem[Xia and Bouganis(2022{\natexlab{a}})]{Xia2022OnTU}
Guoxuan Xia and Christos-Savvas Bouganis.
\newblock On the usefulness of deep ensemble diversity for out-of-distribution
  detection.
\newblock \emph{ArXiv}, abs/2207.07517, 2022{\natexlab{a}}.

\bibitem[Xia and Bouganis(2022{\natexlab{b}})]{Xia_2022_ACCV}
Guoxuan Xia and Christos-Savvas Bouganis.
\newblock Augmenting softmax information for selective classification with
  out-of-distribution data.
\newblock In \emph{Proceedings of the Asian Conference on Computer Vision
  (ACCV)}, December 2022{\natexlab{b}}.

\bibitem[Xie et~al.(2013)Xie, Xu, and Chuang]{temporalensemble}
Jingjing Xie, Bing Xu, and Zhang Chuang.
\newblock Horizontal and vertical ensemble with deep representation for
  classification.
\newblock In \emph{International Conference on Machine Learning}, 2013.

\bibitem[Yang et~al.(2021)Yang, Zhou, Li, and Liu]{yang2021oodsurvey}
Jingkang Yang, Kaiyang Zhou, Yixuan Li, and Ziwei Liu.
\newblock Generalized out-of-distribution detection: A survey.
\newblock \emph{arXiv preprint arXiv:2110.11334}, 2021.

\bibitem[Yang et~al.(2022)Yang, Wang, Zou, Zhou, Ding, Peng, Wang, Chen, Li,
  Sun, Du, Zhou, Zhang, Hendrycks, Li, and Liu]{yang2022openood}
Jingkang Yang, Pengyun Wang, Dejian Zou, Zitang Zhou, Kunyuan Ding, Wenxuan
  Peng, Haoqi Wang, Guangyao Chen, Bo~Li, Yiyou Sun, Xuefeng Du, Kaiyang Zhou,
  Wayne Zhang, Dan Hendrycks, Yixuan Li, and Ziwei Liu.
\newblock Open{OOD}: Benchmarking generalized out-of-distribution detection.
\newblock In \emph{Thirty-sixth Conference on Neural Information Processing
  Systems Datasets and Benchmarks Track}, 2022.

\end{thebibliography}
\appendix
\onecolumn

\section{Experimental Configuration}\label{app:exp}
\label{asec:experimental config}

This section will provide detailed information about the datasets used for training, development, evaluation and detection. It will also give the exact training and various hyperparameters used for all models.

\subsection{Datasets}

We utilise two training sets WMT16/20, each with a pair of development and evaluation datasets based on newstest13/14 and newstest19/20. Additionally, we utilise three out-of-domain datasets for evaluating detection performance of a wide range of transformer models, see Table \ref{tab:datasets}. As stated previously, all data is cleaned and tokenized using Moses\footnote{\url{github.com/moses-smt/mosesdecoder}}. For WMT16, a shared dictionary is learned using BPE with 32,000 merge operations. On WMT20 we learn disjoint dictionaries using BPE with 40,000 merge operations. A consequence of the larger disjoint dictionary on WMT20 is the significantly lower number of unknown tokens in the OOD datasets.

\begin{table*}[h!]
	\centering{}
	\begin{minipage}[t]{1.0\textwidth}%
		\begin{center}
			\caption{Dataset information together with average source and target sentence sizes post tokenization and processing. The OOD testsets Khresmoi, MTNT and KFTT have two quoted numbers for each field as they were processed using either the En-De WMT16 or En-Ru WMT20 BPE based dictionaries. Additionally, only source side information is provided for OOD sets as these are only used for unsupervised uncertainty estimation.}
			\vspace{-2mm}
			\def\arraystretch{1.08}
			\makebox[\textwidth][c]{
				\begin{tabular}{cc|c|cc|c}
					\toprule
    				\multirow{2}{*}{\textbf{Dataset}} & \multirow{2}{*}{\textbf{Type}} &
    				\textbf{Number of} & \multicolumn{2}{c|}{\textbf{Tokens per Sentence}} & \textbf{Fraction of Unknown} \\
    				& & \textbf{Sentences} & \textbf{Source} & \textbf{Target} & \textbf{Tokens in Source}\\
    				\midrule
    				En-De WMT16 & policy, news, web & 4.5M & 29.5 & 30.6 & 0.01\% \\
    				En-De newstest13 & \multirow{2}{*}{news} & 3.0K & 26.0 & 28.0 & 0.00\% \\
    				En-De newstest14 & & 3.0K & 27.6 & 29.1 & 0.00\% \\
    				\midrule
    				En-Ru WMT20 & policy, news, web & 58.4M & 27.8 & 27.5 & 0.00\% \\
    				En-Ru newstest19 & \multirow{2}{*}{news} & 2.0K & 29.9 & 33.4 & 0.00\% \\
    				En-Ru newstest20 & & 2.0K & 30.9 & 32.5 & 0.00\% \\
    				\midrule
    				Khresmoi & medical & 1.0K & 30.9/30.3 & --- & 0.78\%/0.00\% \\
    				MTNT & noisy reddit & 1.4K & 21.1/21.3 & --- & 0.45\%/0.06\% \\
    				KFTT & encyclopedia & 1.2K & 35.4/35.2 & --- & 1.46\%/0.01\% \\
					\bottomrule
			\end{tabular}}
			\label{tab:datasets}
		\end{center}
	\end{minipage}
\end{table*}

\subsection{En-De WMT16 Training}
\label{assec:wmt16}

We use the base transformer from \citep{aiayn} implemented in \texttt{fairseq} \citep{fairseq} and train it using 4 NVIDIA\textcopyright\hspace{0mm} A100 with an update frequency of 32. This is virtually equivalent to training on $4 \times 32 = 128$ GPUs. A per-gpu batch has a maximum of 3584 tokens. Models are optimized with Adam \citep{adam} using $\beta_1$ = 0.9, $\beta_2$ = 0.98, and $\epsilon$= 1e-8. We use a similar learning rate schedule to \citet{aiayn}, i.e., the learning rate increases linearly for 4000 warmup steps to a learning rate dependent on $d_{\tt model}$ after which it is decayed proportionally to the inverse square root of the number of steps:
\begin{equation*}
    \eta = ({\tt step} \cdot d_{\tt model})^{-0.5} \min \hspace{-0.8mm} \left(1, \frac{{\tt step}}{{\tt warmup}} \right)^{1.5}
\end{equation*}
We use label smoothing with 0.1 weight for the uniform prior distribution over the vocabulary. The last 10 weight checkpoints were averaged. Training was stopped after 31 epochs corresponding to approximately a total of 18 GPU-hours. At inference, a beam of 4 with a length-penalty of 0.6 is used for all models. The Deep Ensemble consists of 5 of such models.

\textbf{KD of Deep Ensemble}: Knowledge distilled models are first initialised by one of the teacher members and then trained using the knowledge distillation loss $\mathcal{L}_{\tt KD}$ provided in Section 2.2 with $\lambda = 0.50$. The student was trained with a warmup of 1026 steps (3 epochs), from $\eta = 4.0 \times 10^{-4}$ to $\eta = 7.0 \times 10^{-4}$ after which it decays for a total of 24 epochs. A temperature of $T = 0.8$ was used in the KL-divergence loss as this was found to be mildly beneficial. All other hyperparameters match the standard case above.

\textbf{Snapshot Ensemble}: The Snapshot Ensemble was generated by first starting from the last checkpoint of a standard trained transformer. At this point, a cyclic triangular learning rate schedule \citep{cyclic} was employed oscillating between the values of $\eta_{\tt min} = 1.0 \times 10^{-4}$ and $\eta_{\tt max} = 1.0 \times 10^{-3}$ with a period of 3 epochs. Note that the maximum learning rate in this cyclic phase is notably larger than the peak learning rate ($7.0 \times 10^{-4}$) during standard training  This setting was run for 15 epochs generating an ensemble with 5 members.

\textbf{KD of Snapshot Ensemble}: This system was trained using the same parameters as the Deep Ensemble distilled students but was however, trained for only 12 epochs since it converged faster.

\textbf{EDD \& L-EDD}: All of the EDD and L-EDD systems were distribution distilled from the Snapshot Ensemble using the same setup as "KD of Snapshot Ensemble". We chose $\beta = 0.10$ by evaluating the translation performance of a range of values $\beta \in \{0.05, 0.10, 0.20, 0.50\}$ on the development newstest-13 set, see Section 3.1.

\subsection{En-Ru WMT20 Training}
\label{assec:wmt20}

We use the big transformer from \citet{aiayn} again implemented in \texttt{fairseq} and trained using 4 NVIDIA\textcopyright\hspace{0mm} A100 with an update frequency of 32. A per-gpu batch has a maximum of 5120 tokens. Dropout was set to a value of 0.10 and weight decay to 0.0001. In this case we train the model for 20 epochs, corresponding to 53960 update steps and approximately 230 GPU-hours. The last 5 checkpoints were averaged leading to improved performance. At inference, a beam of 5 with a length-penalty of 1.0 is used for all models.

\textbf{Snapshot Ensemble}: Based on the last checkpoint of a standard trained big transformer, a triangular cyclic learning rate is utilised, oscillating between $\eta = 5.0 \times 10^{-5}$ and $\eta = 5.0 \times 10^{-4}$ every 2 epochs for 10 epochs. This results in an ensemble with 5 members. 

\textbf{KD of Snapshot Ensemble}: Similar to the previous section, the distillation student is initialised from its teacher but is trained using a learning rate warmup of 2698 steps (one epoch) from $\eta = 2.0 \times 10^{-4}$ to $\eta = 4.0 \times 10^{-4}$ after which it decays for a total of 12 epochs. The last 3 or 5 epochs are averaged, based on development newstest19 performance.

\textbf{L-EDD}: Following distillation, L-EDD (Laplace) models are trained using the same parameters. The best-found parameter $\beta = 0.10$ in the WMT'16 experiments is to be used here. No hyperparameter search is performed at this stage.

\end{document}